\documentclass{article}

\usepackage[preprint]{corl_2026} 
\usepackage{xspace}

\usepackage{amsmath}      
\usepackage{amssymb}      
\usepackage{booktabs}     
\usepackage{colortbl}     
\usepackage{subcaption}   
\usepackage{graphicx}     
\usepackage{wrapfig}      
\usepackage{algorithm}    
\usepackage{algpseudocode}
\usepackage[capitalize]{cleveref} 
\usepackage{enumitem}

\newcommand{\cmark}{\checkmark}
\newcommand{\xmark}{$\times$}

\definecolor{seenshade}{rgb}{0.93,0.93,0.93}
\definecolor{todoorange}{rgb}{0.85,0.42,0.10}
\definecolor{projgreen}{rgb}{0.13,0.55,0.13}  

\DeclareRobustCommand{\methodname}{\textsc{ReCAP}\xspace}

\title{Retrieve, Don't Retrain: Extending Vision-Language-Action Models to New Tasks \\at Test Time}

\author{
  Jeongeun Park\\
  NAVER AI Lab\\
  \And
  Juhan Park\\
  Korea University\\
  \And
  Taekyung Kim \\
  NAVER AI Lab\\ 
  \AND
  Sungjoon Choi\\
  Korea University\\
  \And
  Dongyoon Han\\
  NAVER AI Lab\\
  \And
  Sangdoo Yun \\
  NAVER AI Lab\\
}

\begin{document}
\maketitle
\vspace{-2.6em}
\begin{center}
\href{https://recap-robot.github.io/}{\textcolor{projgreen}{\textbf{\texttt{recap-robot.github.io}}}}
\end{center}
\vspace{-0.6em}



\begin{abstract}
Extending a vision-language-action (VLA) policy to a new task typically requires task-specific teleoperated demonstrations and per-task fine-tuning, making adaptation costly in both data collection and compute.
In this paper, we show that this target-side per-task adaptation cost can be replaced by \textbf{retrieval}.
Our retrieval augmented policy is trained once on paired demonstrations from the target embodiment (\textbf{query}) and a cheaper embodiment (\textbf{pool}, \textit{e.g.}, human-hand video), then frozen. 
New tasks are added at deployment by appending pool-side demonstrations to a retrieval pool. 
The frozen policy conditions on retrieved trajectories at every control step, so new tasks are absorbed by indexing data rather than updating parameters. 
Fine-tuning is needed only to take on a new, unseen embodiment, not for each new task.
We show that retrieval improves policies beyond a specific backbone, including standard VLA policies, but its effect is especially pronounced in Cosmos Policy, a video-generation-based world-action model (WAM). In this setting, retrieval supplies coarse task progression, while the WAM’s future-image objective provides an additional visual consistency signal that strengthens the retrieval-conditioned actions.
On PushT, we study how retrieval provides a reusable high-level motion prior for cross-embodiment generalization to unseen goal angles, while on RoboTwin 2.0 our method outperforms cross-embodiment baselines on unseen tasks, and we additionally demonstrate the method on a real robot.
\end{abstract}



\section{Introduction}


General-purpose robot policies~\citep{kim2024openvla,lee2025molmoact,intelligence2025pi,bjorck2025groot,zheng2026xvla,liu2025rdtb, kim2026cosmospolicy, ye2026world} aim to execute open-ended manipulation behaviors from natural-language instructions while generalizing across diverse environments, tasks, and embodiments.
Yet a new embodiment still requires its own teleoperated demonstrations and per-task fine-tuning, so cost grows with each new task added. 
We argue that this per-task cost is avoidable: behavioral knowledge from a cheap, data-rich source (\textit{e.g.}, human-hand video demonstrations) can transfer to the target embodiment through a \textbf{retrieval rather than retraining} paradigm.

The cost of the previous approach is twofold. On the data side, target-embodiment demonstrations would be collected through teleoperation, which is slow, hardware-bound, and roughly $18\times$ slower to acquire than equivalent human-hand demonstrations~\citep{shah2025mimicdroid,wang2023mimicplay}. 
On the compute side, modern vision-language-action (VLA) models and robot foundation models operate over high-dimensional visual and action sequences, so per-task fine-tuning of recent world-action models (WAM)~\cite{kim2026cosmospolicy,yuan2026fastwam,ye2026world} costs roughly $24$ GPU-hours per task and continues to scale with model size, context length, and action horizon. 
Both costs compound with every new task introduced.

We propose \methodname{} (\textbf{Re}trieval-\textbf{C}onditioned \textbf{A}ction \textbf{P}olicy), which shifts adaptation from repeated optimization to retrieval over a reusable pool of source-embodiment demonstrations. The policy is trained \textit{once} to bridge the gap between source and target embodiments and is then frozen; behavioral coverage expands by \textit{simply appending new demonstrations to the retrieval memory}. 

\methodname{} builds on a world-action model (WAM)~\citep{kim2026cosmospolicy,ye2026world,pai2025mimic,yuan2026fastwam,li2026causal,wang2026world,li2025unified}, specifically Cosmos Policy~\citep{kim2026cosmospolicy}. We parameterize the action latents as a \emph{residual} over retrieved trajectories: retrieval supplies the coarse high-level motion and task progression, while the policy learns only the embodiment-specific dynamics needed to execute the behavior on the target robot. 
Crucially, the WAM's future-image prediction objective enforces consistency between the retrieved trajectory and the predicted evolution of the scene, a visual alignment signal that becomes informative \emph{only} when paired with retrieval in unseen tasks, and that we find especially beneficial for long-horizon behaviors where high-level motion structure dominates.



\begin{figure}[t]
    \centering
    \includegraphics[width=0.95\linewidth]{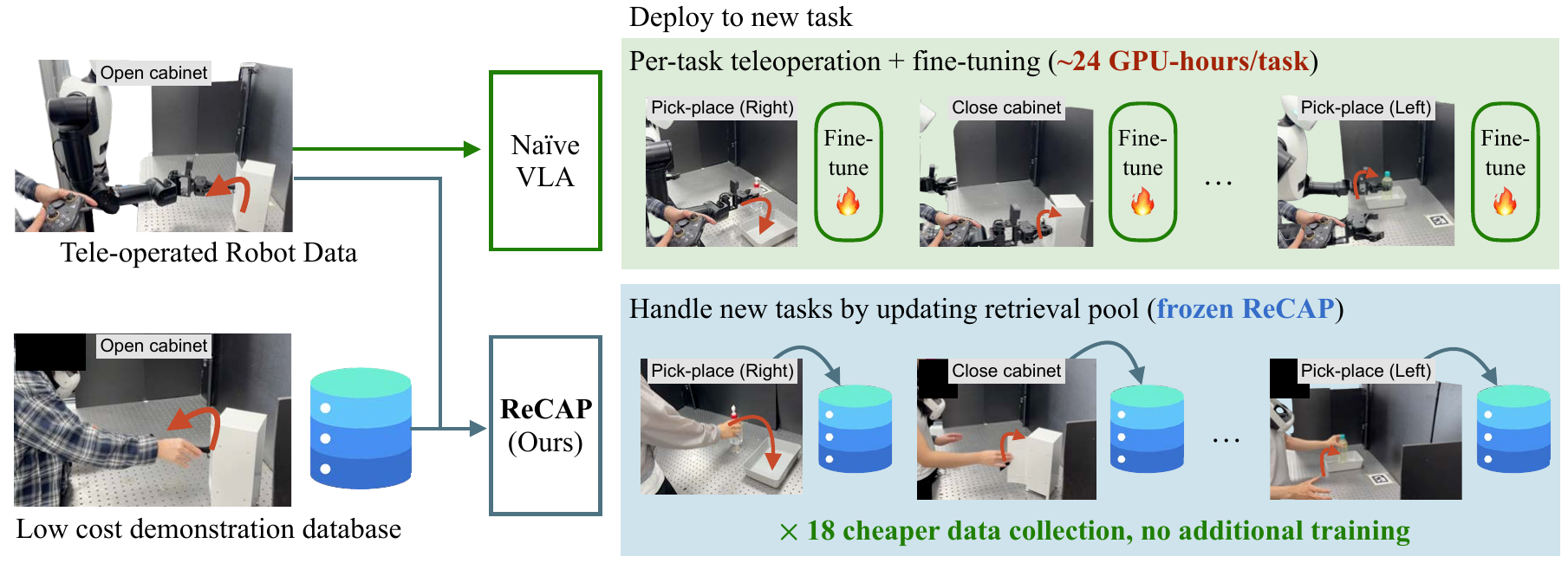}
    \caption{\textbf{\methodname overview.} Instead of teleoperating each new task and fine-tuning the policy (top, $\sim$24 GPU-hours/task for Cosmos Policy~\citep{kim2026cosmospolicy}), ReCAP appends cheap human-hand demonstrations to a retrieval pool while keeping the policy frozen (bottom), 18$\times$ cheaper~\cite{ shah2025mimicdroid, wang2023mimicplay}, no additional training.}
    \label{fig:teaser}
    \vspace{-0.5\baselineskip}
\end{figure}

The main contribution of this paper is threefold: a paradigm that adapts a policy to new tasks entirely at test time, absorbing each new task by extending a retrieval pool with cheap pool-embodiment demonstrations while the policy stays frozen with no parameter updates; a retrieval-conditioned residual policy on a WAM (i.e., Cosmos Policy~\citep{kim2026cosmospolicy}) in which retrieval supplies the high-level motion so the policy learns only the embodiment-specific correction, reinforced by the WAM's future-image objective that is informative only when paired with retrieval; and consistent gains over cross-embodiment baselines on PushT~\cite{chi2025diffusion} ($34.9\%$ vs.\ $6.0\%$ on seven unseen angles) and RoboTwin 2.0~\cite{chen2025robotwin} ($31.5\%$ vs.\ $26.0\%$ on five unseen tasks), with a pool-progression study confirming monotonic coverage growth without parameter updates and a further real-robot validation.

\section{Related Work}

\noindent\textbf{World action models.}
Recent VLA policies inherit either a pretrained language backbone with an added action head (OpenVLA~\citep{kim2024openvla}, $\pi_{0.5}$~\citep{intelligence2025pi}, GR00T~N1.6~\citep{bjorck2025groot}) or a pretrained video model that folds actions into the same generative process (DreamZero~\citep{ye2026world}, Cosmos Policy~\citep{kim2026cosmospolicy}, mimic-video~\citep{pai2025mimic}, Fast-WAM~\citep{yuan2026fastwam}). We call the latter family world-action models (WAMs); their video backbone is pretrained on internet-scale data and already encodes semantics and physical dynamics, so the policy learns only control on top, with action and future-observation prediction emerging from one shared video generation. These WAMs train on target-robot demonstrations alone, leaving their dynamics priors unpaired with cheaper cross-embodiment supervision, which we address by building a WAM on Cosmos Policy that conditions on retrieved pool-embodiment trajectories at training and deployment (\cref{sec:method-wam}).

\noindent\textbf{Retrieval and Cross-embodiment policy transfer.}
Retrieval-based imitation~\citep{du2023behavior} adapts a policy with relevant demonstrations rather than training from scratch on each task. Within a single embodiment, FlowRetrieval~\citep{lin2024flowretrieval} co-trains the policy on optical-flow-retrieved data, and STRAP~\citep{memmel2025strap} trains a task-specific specialist at deployment from DTW-matched sub-trajectories. To cut data cost, attention has turned to cheaper human demonstrations~\citep{punamiya2026egobridge, jain2024vid2robot,bahety2024screwmimic}: \citet{hong2025hand} uses a single human-hand demonstration to retrieve matching robot sub-trajectories and efficiently fine-tunes a policy on them for fast adaptation. These methods still pay for each new task with training, whether by co-training or a test-time fine-tune.
Other methods reduce this training cost by using human data more directly. R+X~\citep{papagiannis2025rx} retrieves everyday human videos and executes them directly without training, but it stays in the human domain and does not learn an embodiment-specific correction. In-context methods such as MimicDroid~\citep{shah2025mimicdroid} condition on a human prompt, which a user must hand-pick for each task. We instead learn the cross-embodiment gap once on paired (query, pool) data and freeze it; the user then grows the retrieval pool with new pool-embodiment demonstrations at test time, which broadens the policy’s task coverage without retraining.

\section{Problem Formulation}
\label{sec:problem}

\noindent\textbf{Setting.}
We study a cross-embodiment imitation setting with two sides: a \emph{query} side that we want to control autonomously at deployment, carrying a \emph{target embodiment} (e.g., a robot arm), and a \emph{pool} side carrying a \emph{pool embodiment} that is easier to collect demonstrations from but is never deployed (e.g., a human hand).
The two embodiments differ in geometry, contacts, and dynamics, and differ sharply in data cost, since a query demonstration requires a teleoperation rig and operator while a pool demonstration only needs a lightweight tracker on a human. Transfer between them rests on two assumptions: a shared state/action representation (we use $\mathrm{SE}(3)$ end-effector pose plus a gripper signal where applicable), and motions that are semantically similar at the trajectory level, so a coarse plan derived from a pool trajectory is informative for the query.

\noindent\textbf{Train and test access.}
At training time we assume that the model is given paired demonstrations $\mathcal{D}_{\text{train}}^{\text{query}}$ and $\mathcal{D}_{\text{train}}^{\text{pool}}$ on a fixed task distribution, each a set of state-action pairs $\{(s_t, a_t)\}$ collected on the corresponding embodiment.
At test time the model faces \emph{new} tasks outside this distribution; only pool demonstrations $\mathcal{D}_{\text{test}}^{\text{pool}}$ are provided, no additional query data is collected, and model parameters are not updated.
The model is rolled out on the target embodiment, so the current query state $s_t^{\text{query}}$ is observed at every control step.

\noindent\textbf{Retrieval-conditioned action prediction.}
Let $\mathcal{D}^{\text{pool}}$ denote the active retrieval pool, $\mathcal{D}_{\text{train}}^{\text{pool}}$ at training and $\mathcal{D}_{\text{test}}^{\text{pool}}$ at deployment. At each step $t$, we select $t' = \arg \min_{t'} d(s_t^{\text{query}}, s_{t'}^{\text{pool}})$ over $\mathcal{D}^{\text{pool}}$, where $d$ is a feature-space distance specified in \cref{sec:method-retrieval}. The retrieved chunk $(s_{t':t'+H}^{\text{pool}}, a_{t':t'+H}^{\text{pool}})$ of action chunk length $H$ steps together with $s_t^{\text{query}}$ is fed to the policy, which predicts the query action chunk $a_{t:t+H}^{\text{query}}$. 
The same rule applies at training and deployment with only $\mathcal{D}^{\text{pool}}$ changing; crucially, $\mathcal{D}_{\text{test}}^{\text{pool}}$ can be extended at any time without touching $\theta$.


\section{Proposed Method}
\label{sec:method}

We propose \methodname to adapt a policy to new tasks at test time without retraining, by retrieving relevant demonstrations rather than updating weights. The intuition is that the target and pool embodiments largely agree on \emph{what} a task requires and differ mainly in \emph{how} to execute it. A retrieved pool trajectory therefore supplies the shared high-level plan cheaply, leaving the policy to learn only the embodiment-specific correction, so adapting to a new task becomes a matter of indexing data rather than updating parameters, much as retrieval-augmented generation externalizes knowledge into a searchable store (\cref{fig:method-overview}).
\cref{sec:method-wam} details the backbone and its retrieval conditioning, and \cref{sec:method-retrieval} specifies how the retrieved chunk is selected and extended at test time.


\subsection{Retrieval-Augmented World Action Model}
\label{sec:method-wam}

Because the policy stays frozen and conditions on an external pool, new tasks can be absorbed at test time by extending that pool rather than by retraining; we expect coverage to grow with the pool, 
\begin{wrapfigure}{r}{0.55\linewidth}
    \vspace{-1.0\baselineskip}
    \centering
    \includegraphics[width=\linewidth]{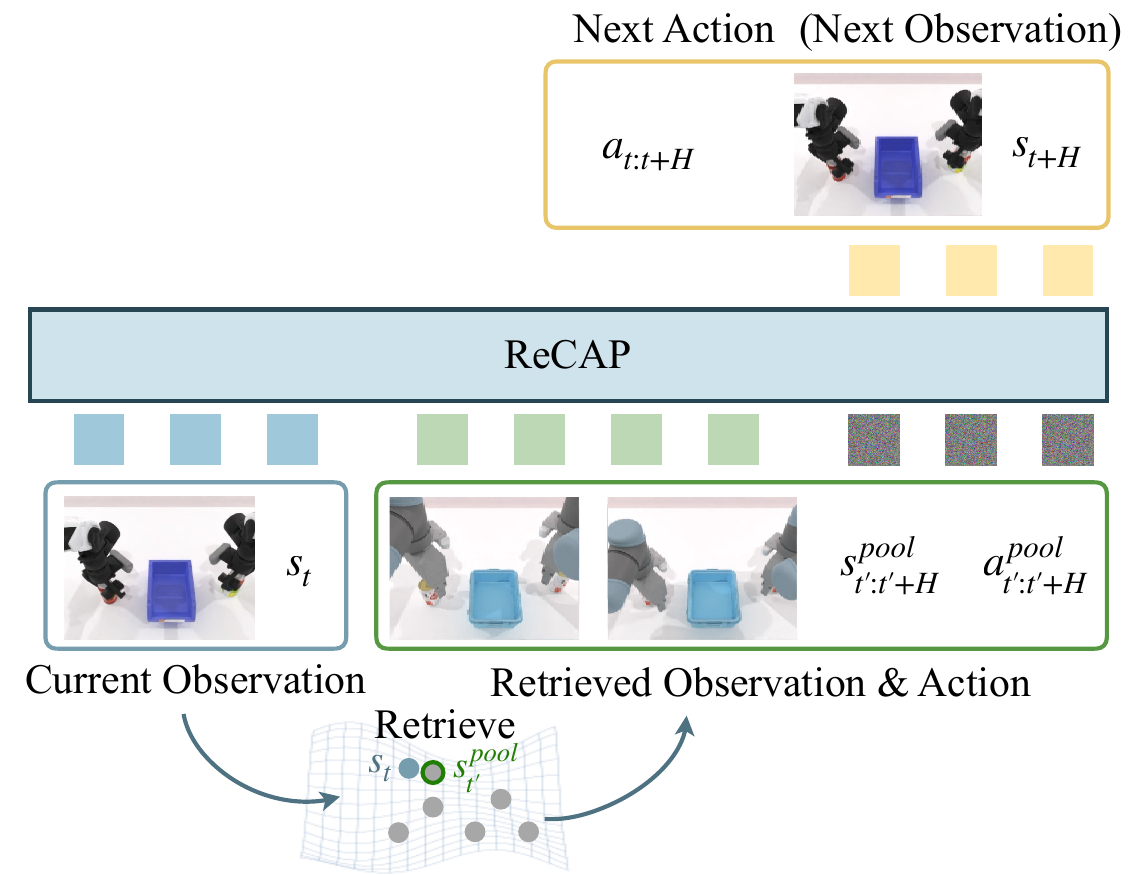}
    \caption{\textbf{\methodname framework.} The current observation retrieves a matching state-action chunk from the pool database; the retrieved chunk and the current observation then condition a world action model that denoises the next action and next observation in one video sequence.}
    \label{fig:method-overview}
    \vspace{-4.0\baselineskip}
\end{wrapfigure}
and cheap pool-embodiment data, such as human-hand video, to partly substitute for target-robot teleoperation.

\noindent\textbf{Backbone and retrieval-conditioned input.}
The backbone is the Cosmos Policy formulation~\citep{kim2026cosmospolicy}, which emits query-side actions and future image observations as one denoised video sequence. 

We extend its conditioning with the retrieved pool-embodiment chunk:
\begin{multline}
    \pi_\theta\!\left(s_t^{\text{query}},\, \bigl(s_{t':t'+H}^{\text{pool}},\; a_{t':t'+H}^{\text{pool}}\bigr)\right) \\
    \longmapsto\; \hat{a}_{t:t+H}^{\text{query}}\,,\; \hat{s}^{\text{query}}_{t+H}.
    \label{eq:policy}
\end{multline}

The retrieved chunk and the observed query frame are encoded into clean latent frames and prepended along the temporal axis as conditioning, while the query-side future actions and observations are denoised from noise (\cref{fig:method-overview}); the language instruction enters via cross-attention. The retrieved chunk thus extends the standard I2V conditioning, a single clean frame, to a clean state-action sub-sequence, with no architectural modification.

\noindent\textbf{Joint training objective (World action model).}
Action and future-image latents are supervised jointly with a single flow-matching loss:
\begin{equation}
    \mathcal{L}(\theta) \;=\; \lambda\, \mathcal{L}_{\text{act}}\!\bigl(\hat a_{t:t+H}^{\text{query}},\, a_{t:t+H}^{\text{query}}\bigr)
    \;+\; \mathcal{L}_{\text{state}}\!\bigl(\hat s_{t+H}^{\text{query}},\, s_{t+H}^{\text{query}}\bigr)
    \label{eq:loss}
\end{equation}
Joint training yields actions aligned with the predicted next state, producing more grounded action outputs. For a standard VLA, we lack $\mathcal{L}_{\text{state}}$ and $\hat{s}^{\text{query}}_{t+H}$.

\noindent\textbf{Residual action parameterization.}
Because the retrieved pool action chunk $a_{t':t'+H}^{\text{pool}}$ already encodes a coarse motion the target should execute, we let the action latents represent only the embodiment-specific correction~\citep{sha2026efficient, schaff2020residual} on top:
\begin{equation}
    \hat a_{t:t+H}^{\text{query}} \;=\; a_{t':t'+H}^{\text{pool}} \;+\; \Delta a_{t:t+H}.
    \label{eq:residual}
\end{equation}
This narrows what the action latents must encode to ``how the query action differs from the pool's,'' the variation actually caused by the embodiment gap. This variation is weakly reflected in action labels but clearly visible in pixels, e.g., how contact happens, how the gripper closes. Residual focuses the action latents on this correction, and state prediction provides the dense visual signal to learn it.

\subsection{Retrieval}
\label{sec:method-retrieval}
At each control step $t$, the policy retrieves a pool-embodiment index $t'$ from $\mathcal{D}^{\text{pool}}$ whose surrounding chunk best matches the current query context. We first form a candidate set $\mathcal{C}^{\text{traj}}_t$ by taking the top-$K$ trajectories closest to the query under a composite initial-frame descriptor $\psi_0$, a language embedding of the goal, initial task-relevant object positions (via SAM 3~\citep{carion2025sam3segmentconcepts}), and initial proprioception. Within $\mathcal{C}^{\text{traj}}_t$, the index distance $d$ in \cref{sec:problem} is a weighted sum of $L_2$ distances over object pose, proprioception, and the upcoming action chunk (training only; dropped at inference), and a cosine distance over a DINOv3~\citep{simeoni2025dinov3} image feature.

At inference, new pool-embodiment demonstrations $\mathcal{D}_{\text{test}}^{\text{pool}}$ replace the active pool and are reindexed under $\psi_0$ and the features above; retrieval re-runs every step, so $\mathcal{D}_{\text{test}}^{\text{pool}}$ can grow within a session.

\section{Experiments}
We evaluate the proposed \methodname policy in three challenging cross-embodiment settings. 
PushT variant~\citep{chi2025diffusion} (\S\ref{sec:pusht-results}) provides a controlled environment for analyzing how retrieval improves generalization and why retrieval-conditioned WAMs are effective. 
RoboTwin 2.0~\cite{chen2025robotwin} (\S\ref{sec:robotwin}) evaluates whether the same paradigm scales to multi-task dual-arm manipulation, while real-robot experiments (\S\ref{sec:realrobot}) test whether unseen tasks can be absorbed through retrieval alone using human-hand demonstrations without additional robot training. Across all setups, we study the same hypothesis: retrieval supplies coarse task progression, while the policy learns only the embodiment-specific dynamics needed to execute the behavior on the target robot.

\subsection{Experiment Setup}
\begin{figure}[t]
    \centering
    \includegraphics[width=\linewidth]{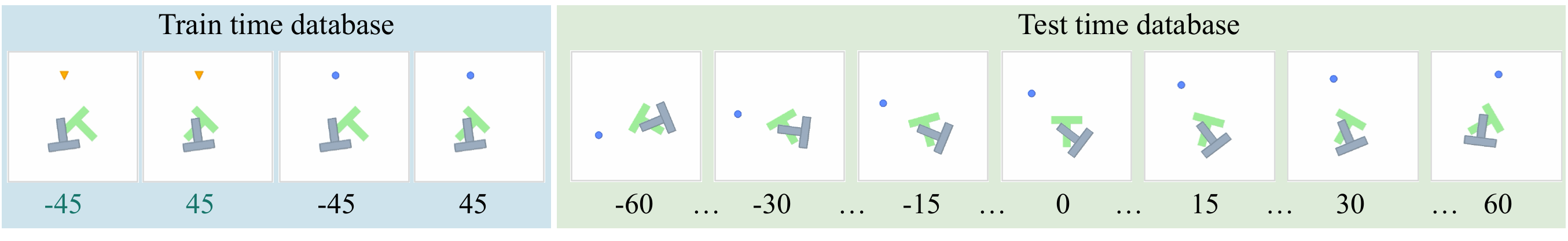}
    \caption{\textbf{PushT cross-embodiment pool database setting.} The training set pairs the \emph{triangle} (target) and \emph{disc} (pool) at $\pm45^\circ$. The test set is a pool database of disc-pusher demonstrations spanning all goal angles, which the frozen triangle policy retrieves from on the seven unseen angles.}
    \label{fig:settings}
    \vspace{-0.5\baselineskip}
\end{figure}

\noindent\textbf{PushT Environment.}
In the 2D PushT benchmark~\citep{chi2025diffusion, cadene2024lerobot} an agent pushes a T-shaped block to a goal pose; we make it cross-embodiment with two pushers of different contact dynamics, a \emph{triangle} (target) and a \emph{disc} (pool), and take the goal rotation angle as the task axis. Training uses $100$ paired \{\emph{triangle}, \emph{disc}\} demonstrations at $\pm45^\circ$; the triangle is then evaluated on nine different angles ranging from $-60^\circ$ to $+60^\circ$ in $15^\circ$ steps, seven of them unseen. The test-time pool holds disc-pusher demonstrations at $5^\circ$ resolution over $[-60^\circ, +60^\circ]$, added without retraining.

\noindent\textbf{RoboTwin Simulation Environment.}
On RoboTwin 2.0~\citep{chen2025robotwin}, we take Aloha-Agilex~\citep{zhao2023learning} as the target and UR5 as the pool, training on five paired \{target, pool\} tasks and evaluating on five unseen ones (\cref{tab:robotwin-baselines}). A test-time pool-progression eval grows the pool through five levels (i.e., $11$, $17$, $23$, $29$, $35$ tasks), each a strict superset of the previous, with the policy frozen throughout.

\noindent\textbf{Real Robot Setup.}
On a physical robot, the pool embodiment is a human hand (video with wrist-pose tracked in VR) and the target is the teleoperated robot. We fine-tune on a single task, \textit{open-cabinet}, with $25$ paired demonstrations. 
Then we freeze the policy and evaluate on three tasks: the seen \textit{open-cabinet} and two held out from fine-tuning, \textit{place-bottle-in-plastic-box} and \textit{close-cabinet} (\cref{fig:realrobot-setup}). The only test-time exposure to the held-out behaviors is $10$ human-hand demonstrations per task added to the pool.

\subsection{PushT Experiments}
\label{sec:pusht-results}

In this section, we study the following aspects of retrieval-conditioned generalization through PushT experiments~\cite{chi2025diffusion}: 
(1) whether expanding the retrieval pool at test time improves unseen-task coverage without retraining, 
(2) whether retrieval benefits more from a WAM backbone than from an action-only policy, and 
(3) whether the retrieved trajectory acts as a reusable high-level motion prior that the policy adapts to the target embodiment.
PushT is a controlled testbed whose generalization axis (i.e., the goal angle) is one-dimensional and densely measurable, which enables exploring these questions.



\begin{figure}[!t]
    \centering
    \includegraphics[width=0.99\linewidth]{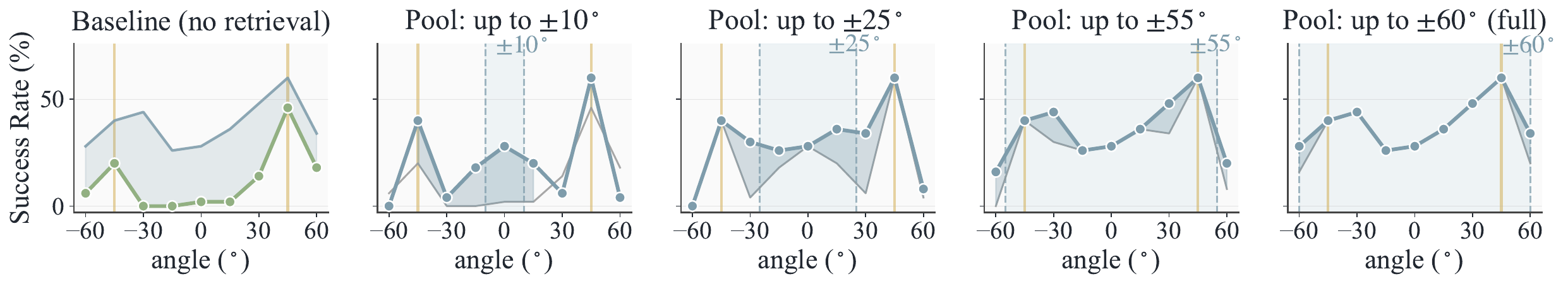}
    \caption{
    \textbf{Test-time pool progression on PushT.} 
    The leftmost panel is the no-retrieval baseline with our full-pool curve overlaid (shaded gap). The other panels show per-angle success as the pool grows with no retraining, with the previous snapshot in gray and the incremental gain shaded.}
    \label{fig:pusht-progression}
\end{figure}

\noindent\textbf{Test-time pool progression.}
Expanding the retrieval pool at test time without parameter updates steadily recovers the unseen angles. We grow the pool with disc-pusher demonstrations at intermediate goal angles and track per-angle success across five snapshots (\cref{fig:pusht-progression}).
Specifically, the unseen-angle average rises monotonically from $6.0\%$ without retrieval to $34.9\%$ at the full pool. Notably, each angle reaches much of its final success \emph{before} its matching pool angle is added, so the policy interpolates over neighboring demonstrations rather than memorizing the nearest one.

\begin{figure}[t]
    \centering
    \begin{subfigure}[b]{0.285\linewidth}
        \centering
        \includegraphics[width=\linewidth]{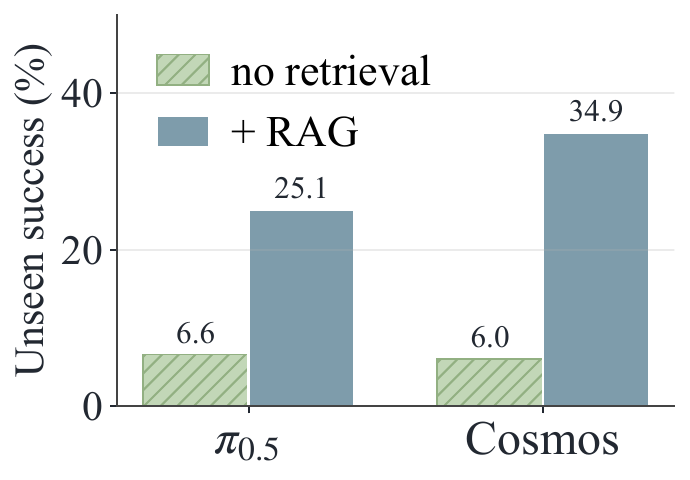}
        \caption{Backbone comparison.}
        \label{fig:pusht-pi05}
    \end{subfigure}
    \hfill
    \begin{subfigure}[b]{0.285\linewidth}
        \centering
        \includegraphics[width=\linewidth]{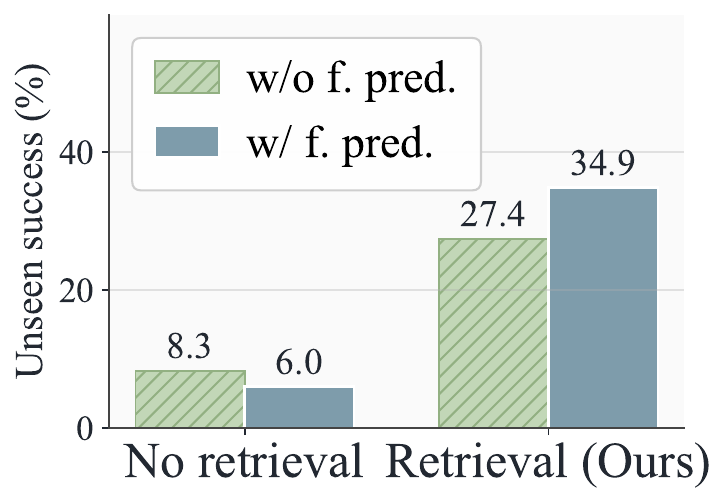}
        \caption{Joint training (Cosmos).}
        \label{fig:pusht-ablation}
    \end{subfigure}
    \hfill
    \begin{subfigure}[b]{0.352\linewidth}
        \centering
        \includegraphics[width=\linewidth]{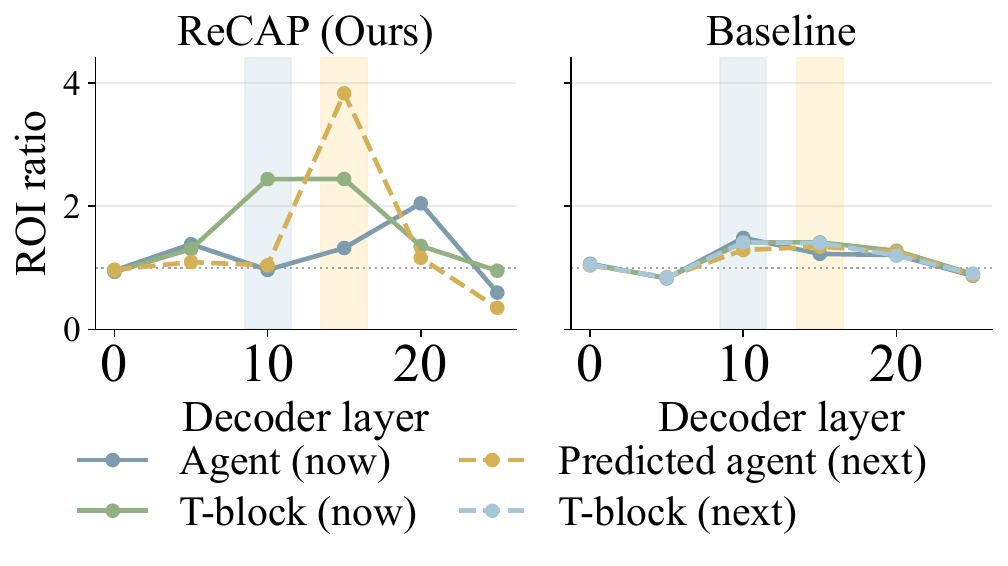}
        \caption{ROI ratio across layers.}
        \label{fig:pusht-mechanism-roi}
    \end{subfigure}
    \caption{\textbf{Comparative analyses of \methodname and baseline on PushT.} (a) Unseen-angle success with and without retrieval on a $\pi_{0.5}$ and a Cosmos (WAM) backbone; retrieval helps both, and the WAM benefits more. (b) The future-image objective improves unseen success only when paired with retrieval. (c) Action-slot attention across decoder layers, which peaks on the T-block and then on the predicted next position under retrieval but stays near uniform without it.}
    \label{fig:pusht-analysis}
    \vspace{-0.5\baselineskip}
\end{figure}

\noindent\textbf{Retrieval and the role of the WAM objective.}
If retrieval already supplies the coarse motion plan, then the remaining learning problem is primarily adapting that plan to the target embodiment dynamics. We therefore hypothesize that a WAM should benefit more from retrieval than an action-only policy, since its future-image objective encourages the retrieved trajectory to remain consistent with the predicted future scene, providing a stronger learning signal for the embodiment-specific dynamics adaptation.
Figure~\ref{fig:pusht-pi05} reveals that retrieval improves both backbones, raising the unseen-angle success of the action-only $\pi_{0.5}$~\citep{intelligence2025pi} from $6.6\%$ to $25.1\%$. 
However, the gain is larger with a WAM backbone.
In \cref{fig:pusht-ablation}, removing the image-prediction objective reduces our model to $27.4\%$, comparable to retrieval-augmented $\pi_{0.5}$ ($25.1\%$), while restoring it improves performance to $34.9\%$.



\noindent\textbf{How the retrieval-conditioned policy works.}
The retrieved chunk acts as a coarse motion prior, while the policy adapts it to the target embodiment rather than planning from scratch. Decoder cross-attention analysis in \cref{fig:pusht-mechanism-roi} reveals a two-stage routing behavior: early layers attend to the manipulated object and retrieved trajectory, whereas later layers shift attention toward the policy's own predicted next state, adapting the retrieved plan to the target embodiment dynamics. 
This structure does not emerge without retrieval, where the ROI ratio remains near $1.0$ across layers, indicating near-uniform attention. Masking the action-to-retrieval attention further confirms that the retrieved trajectory causally influences the generated actions.
\subsection{RoboTwin Simulation Experiments}
\label{sec:robotwin}
In this section, we evaluate the proposed method in a multi-task, dual-arm manipulation simulation environment, RoboTwin 2.0~\citep{chen2025robotwin}. We compare against standard cross-embodiment baselines on seen and held-out unseen tasks, then test whether the test-time pool growth seen on PushT carries over to this multi-task regime.

\begin{table}[t]
    \centering
    \caption{
    \textbf{Quantitative analysis on RoboTwin.} 
    We report per-task success rate ($\%$) on RoboTwin, with Aloha-Agilex as the target embodiment and UR5 as the retrieval pool. 
    The left block shows seen tasks, and the right block shows unseen tasks.
    }
    \label{tab:robotwin-baselines}
    \small
    \setlength{\tabcolsep}{4pt}
    \begin{tabular}{l ccccc c ccccc c}
        \toprule
        & \multicolumn{6}{c}{\textit{Seen tasks}} & \multicolumn{6}{c}{\textit{Unseen tasks}} \\
        \cmidrule(lr){2-7} \cmidrule(lr){8-13}
        Method & PCB & OM & DB & MP & GR & Avg $\uparrow$ & MPP & PBS & CB & HM & LP & Avg $\uparrow$ \\
        \midrule
        Baseline~\citep{kim2026cosmospolicy} & 47.5 & 10.0 & 25.0 & 30.0 & 50.0 & 32.5 &  0.0 &  0.0 & 20.0 &  0.0 &  0.0 &  4.0 \\
        Retrieval Only            & 30.0 &  7.5 & 22.5 &  7.5 & 60.0 & 25.5 &  0.0 & 10.0 & 42.5 & 37.5 & 40.0 & 26.0 \\
        Co-training       &  0.0 &  5.0 &  7.5 & 50.0 & 72.5 & 27.0 &  0.0 &  0.0 & 40.0 &  0.0 & 10.0 & 10.0 \\
         \textbf{\methodname (Ours)}             & \textbf{60.0} & \textbf{12.5} & \textbf{40.0} & 35.0 & 70.0 & \textbf{43.5} & \textbf{5.0} & \textbf{12.5} & \textbf{47.5} & \textbf{47.5} & \textbf{45.0} & \textbf{31.5} \\
        \bottomrule
    \end{tabular}
    \vspace{1em}
    \scriptsize
    \begin{tabular}{@{}lllll@{}}
    PCB = Place Cans Plasticbox & OM = Open Microwave  & DB = Pick Dual Bottles & MP = Move Can Pot   & GR = Grab Roller \\
    MPP = Move Pillbottle Pad    & PBS = Place Bread Skillet        & CB = Click Bell        & HM = Hand-over Mic  & LP = Lift Pot \\
    \end{tabular}
    \vspace{-2.0\baselineskip}
\end{table}

\begin{figure}[t]
    \centering
    \includegraphics[width=0.95\linewidth]{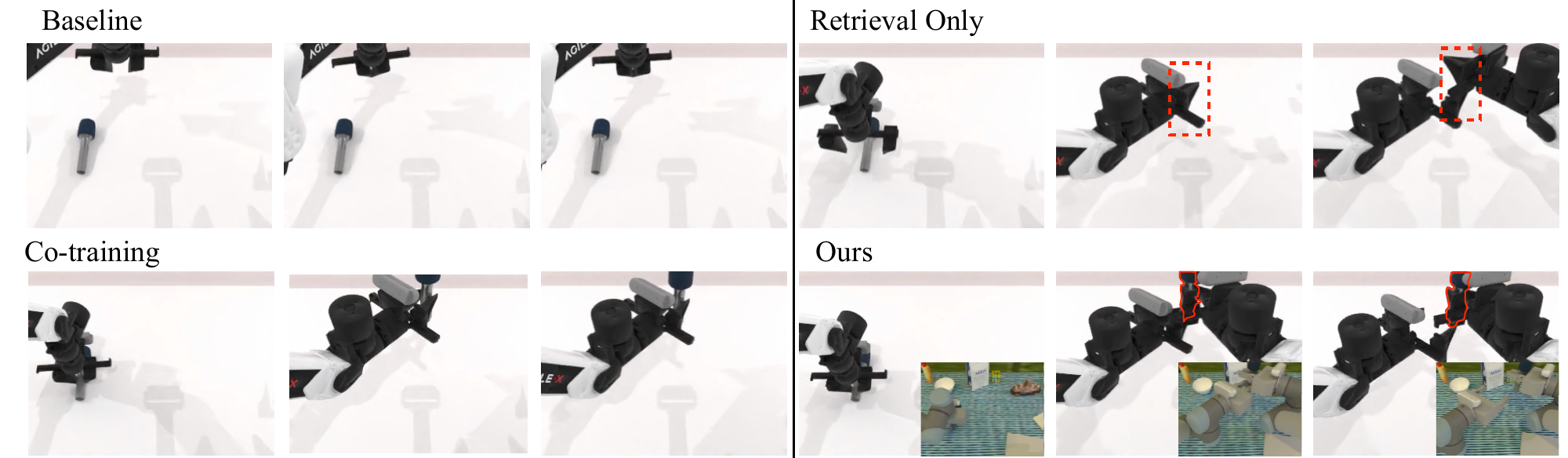}
    \caption{
    \textbf{Qualitative comparison on the held-out hand-over-mic task.} 
    Baseline (top-left) and Co-training (bottom-left) fail to grasp the microphone; Retrieval Only (top-right) knocks it over (red box); Ours (bottom-right) grasps it successfully. Each inset shows the retrieved UR5 chunk that the policy conditions on.
    }
    \label{fig:robotwin-example}
    \vspace{-0.5\baselineskip}
\end{figure}

\noindent\textbf{Baseline comparisons.}
We compare against three cross-embodiment baselines that share our backbone but differ in how they incorporate the UR5 (pool) data. \emph{Baseline}, Cosmos Policy~\citep{kim2026cosmospolicy} is trained on Aloha-Agilex (target) demonstrations alone, with no access to the pool. \emph{Retrieval Only} executes the action sequence of the nearest pool demonstration without learning. \emph{Co-training} is a common cross-embodiment baseline, also used in EgoBridge~\citep{punamiya2026egobridge} and STRAP~\citep{memmel2025strap}, that jointly trains a single policy on the union of target and pool trajectories. 
\cref{tab:robotwin-baselines} shows that \methodname leads on both splits, at $43.5\%$ seen and $31.5\%$ unseen versus $32.5\%$ and $26.0\%$ for the strongest baseline. Replaying the nearest pool trajectory (i.e., Retrieval Only) is competitive only where that trajectory already approximates the target action.
Otherwise, the learned residual on top of retrieval is what closes the gap.
As \cref{fig:robotwin-example} illustrates, the nearest UR5 trajectory collides and dislodges the object, while our policy produces the finer grip orientation the task needs.

\begin{wrapfigure}{r}{0.28\linewidth}
    \vspace{-1.0\baselineskip}
    \centering
    \includegraphics[width=\linewidth]{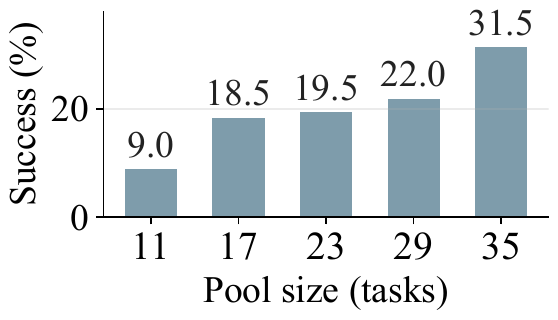}
    \caption{Test-time pool progression on RoboTwin.}
    \label{fig:robotwin-progression}
    \vspace{-1.4\baselineskip}
\end{wrapfigure}
\noindent\textbf{Test-time pool progression.}
Figure~\ref{fig:robotwin-progression} shows that growing the retrieval pool at test time with the policy frozen raises unseen-task success monotonically, from $9.0\%$ to $31.5\%$ at the full pool. Each increase coincides with a held-out task becoming retrievable, and once all five are in the pool, the frozen policy matches its supervised unseen-task average. Cheap pool-embodiment data at deployment can therefore stand in for new target-embodiment demonstrations on tasks unseen during fine-tuning.

\subsection{Real Robot Experiments}
\label{sec:realrobot}

\begin{figure}[t]
    \centering
    \begin{subfigure}[b]{0.7\linewidth}
        \centering
        \includegraphics[width=\linewidth]{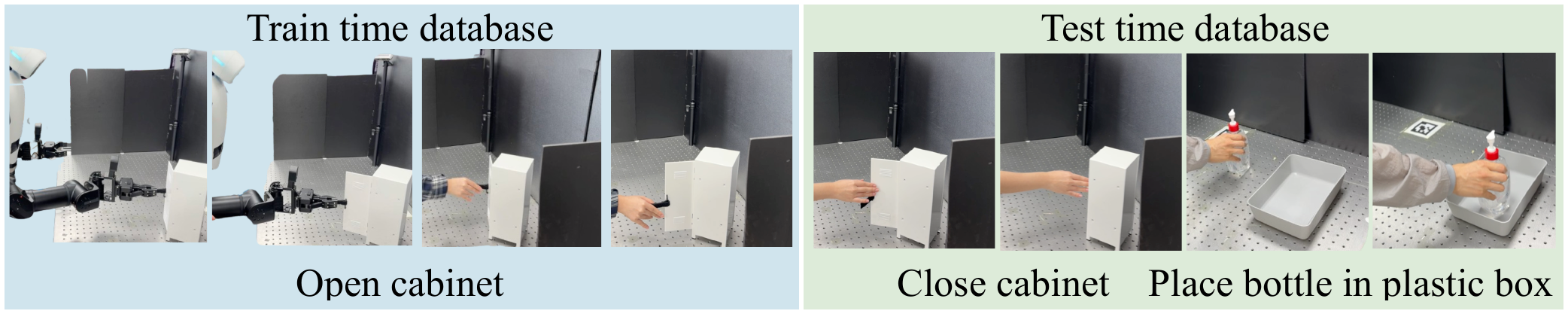}
        \caption{Query, Pool database setting.}
        \label{fig:realrobot-setup}
    \end{subfigure}
    \hfill
    \begin{subfigure}[b]{0.29\linewidth}
        \centering
        \includegraphics[width=\linewidth]{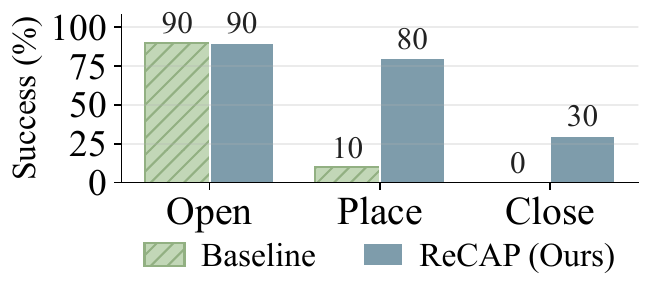}
        \caption{Success rate.}
        \label{fig:realrobot-success}
    \end{subfigure}
    \caption{\textbf{Real-robot experiment.} (a) The training-time database pairs the robot (query) with a human-hand pool; the test-time database adds human-hand demonstrations for the held-out tasks. (b) Per-task success rate over $10$ rollouts, Baseline vs \methodname (Ours).}
    \label{fig:realrobot}
\end{figure}
\begin{figure}[t]
    \centering
    \includegraphics[width=\linewidth]{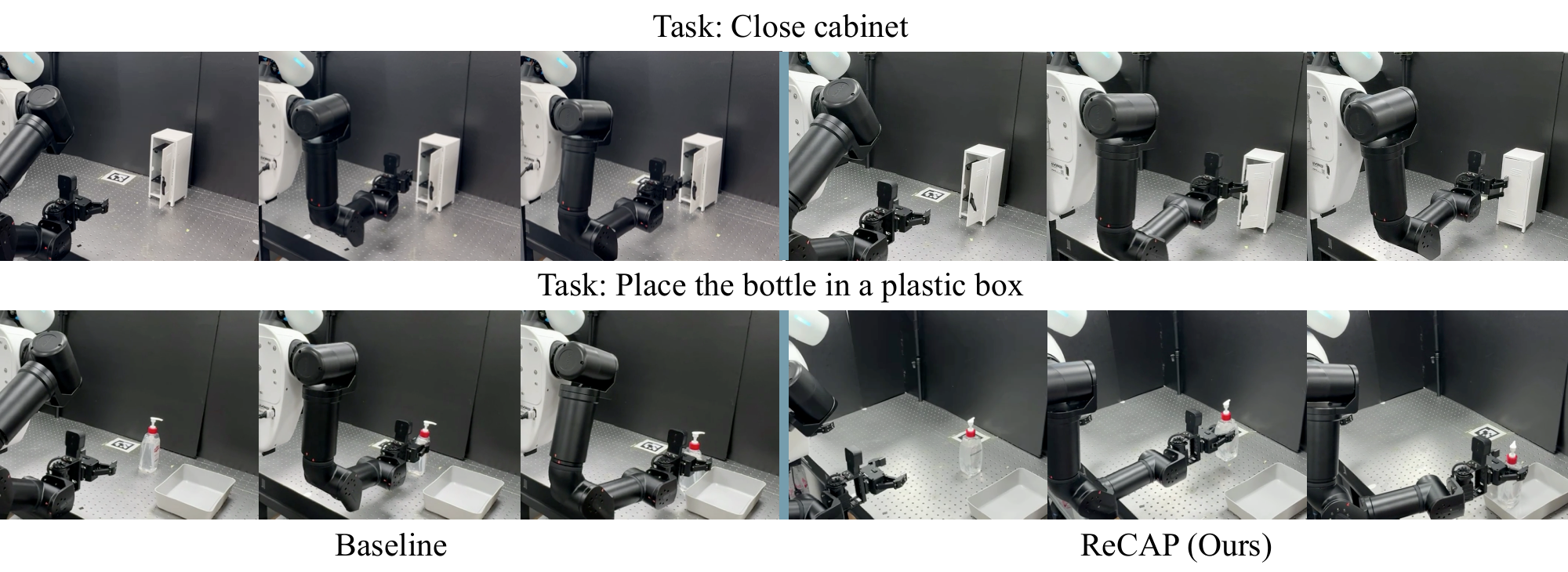}
    \caption{
    \textbf{Real-robot generalization to unseen tasks.} 
    We show rollouts on the two held-out tasks (three frames each; baseline left, ours right). Trained only on \textit{open-cabinet}, the baseline replays that trajectory and fails to close the cabinet (top) or grasp the bottle (bottom), whereas our policy follows the commanded behavior by conditioning on retrieved human-hand chunks.
    }
    \label{fig:realrobot-rollouts}
    \vspace{-0.5\baselineskip}
\end{figure}

We test whether the protocol transfers to real-world robots, despite the large embodiment gap between the human-hand demonstrations in the retrieval pool and the target robot (\cref{fig:realrobot-setup}). On two held-out tasks, the no-retrieval baseline collapses to the trained open-cabinet motion regardless of the target task, reaching only 10\% and 0\%, while retrieval enables the frozen policy to follow the conditioned human trajectory and reach 80\% and 30\% on placing the bottle and closing the cabinet (\cref{fig:realrobot-success}). This indicates that human-hand demonstrations in the pool can partly substitute for additional target-robot teleoperation, even across a substantial embodiment gap. The qualitative results are shown in Figure~\ref{fig:realrobot-rollouts}. 

\section{Discussions}

\noindent\textbf{Summary.}
We extend a world-action-model policy to new tasks without retraining. Trained once to condition on a retrieval pool and predict an embodiment-specific residual on the retrieved trajectory, the frozen policy absorbs a new task by adding cheap pool-embodiment demonstrations at deployment. Across a cross-embodiment PushT variant, RoboTwin~\citep{chen2025robotwin}, and a physical robot, this improves generalization to unseen angles and tasks over cross-embodiment baselines, with success growing as the pool expands, and our analysis ties the gain to the WAM's future-image objective acting together with retrieval. Indexing cheaper pool-embodiment data at deployment can thus stand in for collecting new target-robot demonstrations. 

\noindent\textbf{Limitations and Future Work.}
Several constraints point to future work. The target and pool embodiments must share an end-effector action space, since the residual refines a retrieved chunk in a common low-level representation; structurally different action spaces (e.g., a dexterous hand versus a parallel gripper) would require an embodiment-agnostic interface or learned action translator. The pool must also contain trajectories rather than video alone, so video-only sources such as raw human or web video would first need to be lifted into a state-action representation. What constitutes an effective representation for cross-embodiment retrieval also remains an open question; our current descriptor combines language, object pose, proprioception, and visual features, but more scalable or embodiment-invariant representations may further improve transfer.

In addition, the residual formulation becomes less reliable when retrieved motions differ substantially in execution speed or temporal scale, particularly for larger chunks where errors can accumulate over time. Developing retrieval and adaptation mechanisms that remain robust under significant temporal or dynamical mismatch is an important direction for future work. Finally, scaling retrieval beyond curated trajectory pools to in-the-wild video sources such as raw YouTube video remains a promising step toward broadly reusable robot experience.

\clearpage


\bibliography{example}  

\clearpage
{\LARGE\bfseries Appendix}\\[0.4em]
\appendix

This appendix collects the additional results, analyses, and implementation details that support the main paper. \cref{app:pusht-extra-ablation} reports the full PushT baseline comparison across all nine goal angles and the complete $2\times2$ action-parameterization $\times$ next-state ablation. \cref{app:pusht-mechanism} expands the decoder analysis of \cref{sec:pusht-results}, giving the full attention probe, the no-retrieval comparison, and the causal masking interventions that the main text only summarizes. \cref{app:pusht-failure} builds on that same analysis to examine the failure rollouts, separating them from matched successes and relating the attention differences to behavioral failure clusters. \cref{app:robotwin-setup} details the RoboTwin 2.0 train and test tasks, the excluded query episodes, and the progressive retrieval pool. \cref{app:retrieval-detail} spells out the full two-stage retrieval rule, trajectory prefilter then subframe scoring at training and inference, that the main text abbreviates. \cref{app:hyperparams} lists all backbone, retrieval, and training hyperparameters.

\section{Additional PushT Results}
\label{app:pusht-extra-ablation}

\begin{table}[h]
    \centering
    \caption{\textbf{PushT generalization across goal angles (success rate, \%, higher is better).} Training covers only $\pm45^\circ$ (shaded Seen columns); the other seven angles are unseen at training time. Bold marks the per-column best.}
    \label{tab:pusht-baselines}
    \small
    \setlength{\tabcolsep}{4pt}
    \newcommand{\seen}[1]{\cellcolor{seenshade}#1}
    \begin{tabular}{lccccccccccc}
        \toprule
                                 & \multicolumn{9}{c}{Goal angle ($^\circ$)} & \multicolumn{2}{c}{Avg $\uparrow$} \\
        \cmidrule(lr){2-10} \cmidrule(lr){11-12}
        Method & $-60$ & \seen{$-45$} & $-30$ & $-15$ & $0$ & $+15$ & $+30$ & \seen{$+45$} & $+60$ & Seen & Unseen \\
        \midrule
        Cosmos Policy             &  6.0 & \seen{20.0} &  0.0 &  0.0 &  2.0 &  2.0 & 14.0 & \seen{46.0} & 18.0 & 33.0 &  6.0 \\
        Retrieval Only              & 24.0 & \seen{10.0} & 16.0 & 10.0 & 14.0 & 12.0 & 18.0 & \seen{22.0} & 24.0 & 16.0 & 16.9 \\
        Co-train (all)            & 16.0 & \seen{22.0} & 20.0 & 18.0 & 20.0 & 20.0 & 20.0 & \seen{24.0} & 20.0 & 23.0 & 19.1 \\
        \midrule
        \textbf{Ours}             & \textbf{28.0} & \seen{\textbf{40.0}} & \textbf{44.0} & \textbf{26.0} & \textbf{28.0} & \textbf{36.0} & \textbf{48.0} & \seen{\textbf{60.0}} & \textbf{34.0} & \textbf{50.0} & \textbf{34.9} \\
        \bottomrule
    \end{tabular}
\end{table}

\paragraph{Comparison with prior cross-embodiment recipes.}
\label{app:pusht-baselines}
\cref{tab:pusht-baselines} reports a comparison against three cross-embodiment baselines that share our backbone but differ in how they incorporate the disc-pusher (pool) data. \textbf{Cosmos Policy} is trained on the triangle (target) demonstrations alone, with no access to the pool. \textbf{Retrieval Only} executes the action sequence of the nearest pool demonstration without learning. \textbf{Co-train (all)} jointly trains a single policy on the union of target and pool trajectories. At the full retrieval pool, our method generalizes beyond the $\pm 45^\circ$ training band and achieves the best success on both splits, $34.9\%$ on unseen angles and $50.0\%$ on seen. Without retrieval the same backbone (Cosmos Policy) reaches only $6.0\%$ on unseen; treating the pool as undifferentiated training data (Co-train) is the strongest baseline at $19.1\%$ unseen but still well short of ours, indicating that pool data is necessary but not sufficient.

\begin{table}[h]
    \centering
    \caption{\textbf{Full action-parameterization $\times$ next-state ablation on PushT} (avg unseen-angle success, \%). Rows toggle the auxiliary next-state objective; columns toggle the action parameterization. The last row reports the non-retrieval baseline at the same backbone.}
    \label{tab:pusht-full-ablation}
    \small
    \begin{tabular}{cccc}
        \toprule
        Next-state pred. & Absolute & Residual & $\Delta_{\text{Res}-\text{Abs}}$ \\
        \midrule
        \xmark & $26.3$ & $27.4$ & $+1.1$ \\
        \cmark & $27.4$ & $34.9$ & $+7.4$ \\
        \midrule
        $\Delta_{\text{\cmark}-\text{\xmark}}$ & $+1.1$ & $+7.4$ & \\
        \midrule
        \multicolumn{4}{c}{No-retrieval baseline: $8.3 \to 6.0$ \,($\Delta_{\text{\cmark}-\text{\xmark}} = -2.3$)} \\
        \bottomrule
    \end{tabular}
\end{table}

\paragraph{Full action-parameterization $\times$ next-state ablation.}
\cref{fig:pusht-ablation} in the main body ablates the next-state objective while fixing the action parameterization to residual. For completeness, \cref{tab:pusht-full-ablation} reports the full $2\times 2$ ablation crossing action parameterization (absolute vs.\ residual over the retrieved trajectory) with the auxiliary next-state objective, together with the non-retrieval baseline. Residual outperforms absolute at both settings of the next-state objective. Next-state prediction has negligible effect under absolute parameterization ($+1.1$ on unseen) but contributes $+7.4$ under residual; the same auxiliary loss is detrimental in the non-retrieval regime ($-2.3$). Together, these results indicate that the benefit of next-state supervision is specific to the WAM~$\times$~retrieval interaction.

\section{PushT Mechanism Analysis}
\label{app:pusht-mechanism}

A retrieval-conditioned policy turns a retrieved chunk into an action by routing its decoder computation along two attention axes, and this section establishes those axes, contrasts them against the no-retrieval baseline, and verifies them causally. The first axis is \emph{intake} at layer 10 (L10), where the action slot reads in the retrieved task region, namely the current and goal T-block poses carried by the retrieved chunk. The second is \emph{commit} at layer 15 (L15), where the slot turns to the policy's own predicted end-of-chunk pose and settles on the action it will execute. We first define the attention probe and its ROI-ratio metric (\cref{sec:pusht-results} introduced the summary view); we then show that these two peaks are the only structured attention in the decoder (\cref{tab:pusht-mech-slots}), that they disappear when the same backbone is trained without retrieval, and that masking either peak degrades success. \cref{app:pusht-failure} carries the same two axes into the failure regime and shows how each one breaks down.

\paragraph{Probe protocol.}
We localize what the action slot looks at by reading decoder cross-attention and normalizing it against a uniform spatial baseline, so that a value above one means the slot concentrates on a region rather than spreading evenly. We probe the (residual, next-state-on) PushT policy at iteration $7000$ across the nine-angle grid of \cref{tab:pusht-baselines}. For each rollout we record decoder cross-attention from the action slot to four input groups: the primary view, the retrieved chunk frames (its first and last frame, denoted ``ret-1'' and ``ret-end''), the retrieved proprioception, and the retrieved action. Three image regions of interest (ROIs) are defined on the primary view and on each retrieved frame: \textsc{t\_block} (the T-block bounding box at the current step), \textsc{t\_block\_next} (the T-block at the step the retrieved chunk ends at), and \textsc{predicted\_end} (the policy's predicted end-of-chunk pose, projected back into the image). For a given (slot, layer), let $A_p$ denote the softmax-normalized attention weight on image patch $p$ (so $\sum_{p \in I} A_p = 1$), and let $R \subset I$ be the patches inside a given ROI. We define the ROI ratio as the mean attention inside $R$ normalized by the per-patch uniform baseline:
\begin{equation}
    \mathrm{ROI\;ratio} \;=\; \frac{\frac{1}{|R|}\sum_{p \in R} A_p}{\frac{1}{|I|}\sum_{p \in I} A_p} \;=\; \frac{|I|}{|R|}\sum_{p \in R} A_p.
    \label{eq:roi-ratio}
\end{equation}
A ratio of $1$ corresponds to uniform attention; a ratio of $k$ means the action slot attends to that ROI $k$ times more strongly than uniform. The analysis uses three data subsets: (i) $50$ rollouts ($5$ success $+\,5$ failure $\times\,5$ angles) for the ours-vs-baseline comparison, (ii) $70$ rollouts ($5$ success $+\,5$ failure $\times\,7$ angles) for failure analysis, and (iii) $21$ ablation runs ($7$ angles $\times\,3$ interventions).

\begin{table}[t]
    \centering
    \caption{\textbf{Decoder ROI peaks at L10 (intake) and L15 (commit).} Action-slot attention to each input group, pooled over the $50$ unseen-angle rollouts. Ratio $>1$ means concentration above a uniform spatial baseline. ``ret-1'' / ``ret-end'' denote the first and last frame of the retrieved chunk.}
    \label{tab:pusht-mech-slots}
    \small
    \begin{tabular}{llcc}
        \toprule
        Slot & Dominant ROI & Layer & Ratio \\
        \midrule
        Primary view              & T-block            & L10 & $2.1$ \\
        Retrieved frame, ret-1    & T-block            & L10 & $5.1$ \\
        Retrieved frame, ret-end  & T-block (next)     & L10 & $5.4$ \\
        \midrule
        Primary view              & Predicted end      & L15 & $3.9$ \\
        Retrieved frame, ret-end  & Predicted end      & L15 & $3.0$ \\
        Retrieved frame, ret-1    & (release)          & L15 & $\approx 1.0$ \\
        \bottomrule
    \end{tabular}
\end{table}

\begin{figure}[t]
    \centering
    \includegraphics[width=0.95\linewidth]{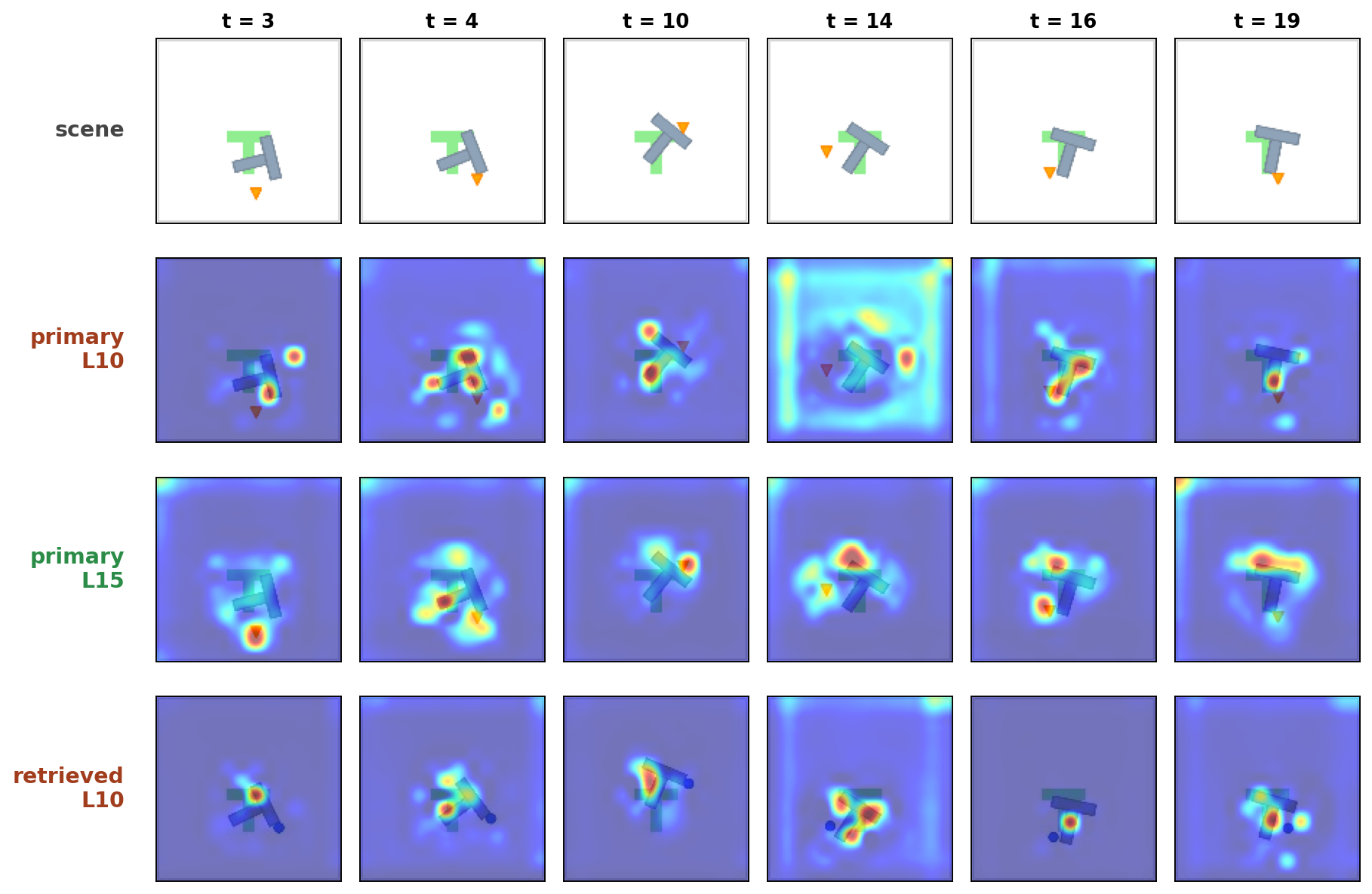}
    \caption{Action-slot attention along a representative rollout. Top row: scene at six timesteps. Subsequent rows: action-slot attention overlaid on the primary view at L10 (intake) and L15 (commit), and on the retrieved frame at L10. The primary L10 row attends to the current T-block; the primary L15 row shifts to the predicted end-of-chunk pose; the retrieved L10 row attends to the retrieved chunk's T-block region, consistent with the per-layer summary in \cref{tab:pusht-mech-slots}.}
    \label{fig:pusht-attn-timeline}
\end{figure}

\paragraph{Two-stage routing: L10 intake, L15 commit.}
The decoder concentrates the action slot's attention at exactly two layers, and these two peaks are precisely the intake and commit axes (\cref{tab:pusht-mech-slots}, traced along a representative rollout in \cref{fig:pusht-attn-timeline}). At L10, the slot attends to the task region carried by the retrieved chunk: it places $5.1\times$ uniform attention on the current T-block in the first retrieved frame (ret-1) and $5.4\times$ on the T-block pose at the end of the retrieved chunk (ret-end). It thus reads the retrieved trajectory temporally, with the earlier frame anchoring where the block is now and the later frame anchoring where the retrieved plan will take it. At L15, the slot drops the retrieved frames and instead concentrates on the policy's own predicted end-of-chunk pose, at $3.9\times$ uniform from the primary view and $3.0\times$ from ret-end, committing to the action it will execute rather than continuing to read evidence. The surrounding layers do no selective routing: encoder-side layers (L0--L5) and late layers (L20--L25) stay near uniform (ratio $\approx 1.0$), so intake and commit are two sharp, well-separated stages rather than a gradual blend. The policy therefore first reads the retrieved plan at L10 and then commits to its own correction of it at L15, which is the decoder-level signature of the residual-over-retrieval behavior.

\begin{table}[t]
    \centering
    \caption{\textbf{Ours vs no-retrieval baseline at the same backbone.} Peak ROI ratios pooled over the five unseen angles common to both runs. The no-retrieval baseline is the Cosmos Policy row of \cref{tab:pusht-baselines}; ROIs, layers, and rollout subsets are defined identically.}
    \label{tab:pusht-mech-baseline}
    \small
    \begin{tabular}{lcc}
        \toprule
        Metric & Ours & Baseline (no retrieval) \\
        \midrule
        L15 predicted-end peak       & $\approx 3.8$ (sharp)                   & $\approx 1.0$ (flat) \\
        L10 task-ROI peak            & $2.0$--$2.5$                            & $1.0$--$1.2$ \\
        L10\,$\to$\,L15 transition   & present                                 & absent \\
        \bottomrule
    \end{tabular}
\end{table}

\paragraph{Both axes require retrieval.}
Neither axis appears when the same backbone is trained without retrieval, so the two-stage routing is a product of retrieval-augmented training rather than of the video backbone itself. Matched on the five unseen angles common to both runs (\cref{tab:pusht-mech-baseline}), the no-retrieval Cosmos Policy keeps every ROI ratio near $1.0$ at every decoder layer: the sharp L15 predicted-end peak ($\approx 3.8$ for ours) flattens to $\approx 1.0$, the L10 task peak ($2.0$--$2.5$) flattens to $1.0$--$1.2$, and the L10$\to$L15 transition is absent altogether. Because the backbone, pretraining, and architecture are identical across the two runs, the only change that introduces the peaks is conditioning on retrieved chunks. The functional contribution of retrieval is therefore not merely to supply extra context but to induce the L15 commitment step in which the policy acts on its own prediction.

\begin{table}[t]
    \centering
    \caption{\textbf{Causal ablation across seven unseen angles.} ``L10 block'' masks the action-to-retrieval cross-attention at decoder layer 10 (similarly for L15); applied separately to the five-success rollouts (5S) and the five-failure rollouts (5F) at each angle. The baseline column reproduces the unseen-angle success of \cref{tab:pusht-baselines}.}
    \label{tab:pusht-mech-ablation}
    \small
    \setlength{\tabcolsep}{4pt}
    \begin{tabular}{lcccc}
        \toprule
        Angle ($^\circ$) & Baseline & L10 block, 5S & L15 block, 5S & L10 block, 5F (rescue) \\
        \midrule
        $-60$ & $28\%$ & $100\!\to\!60\%$ ($-40$) & $100\!\to\!60\%$ ($-40$) & $0\!\to\!\mathbf{40}\%$ \\
        $-30$ & $44\%$ & $100\!\to\!80\%$ ($-20$) & $100\!\to\!80\%$ ($-20$) & $0\!\to\!20\%$ \\
        $-15$ & $26\%$ & $100\!\to\!80\%$ ($-20$) & $100\!\to\!40\%$ ($-60$) & $0\!\to\!20\%$ \\
        $0$   & $28\%$ & $100\!\to\!\mathbf{20}\%$ ($-80$) & $100\!\to\!40\%$ ($-60$) & $0\!\to\!20\%$ \\
        $+15$ & $36\%$ & $100\!\to\!60\%$ ($-40$) & $100\!\to\!80\%$ ($-20$) & $0\!\to\!0\%$ \\
        $+30$ & $48\%$ & $100\!\to\!80\%$ ($-20$) & $100\!\to\!60\%$ ($-40$) & $0\!\to\!20\%$ \\
        $+60$ & $34\%$ & $100\!\to\!\mathbf{40}\%$ ($-60$) & $100\!\to\!40\%$ ($-60$) & $0\!\to\!\mathbf{40}\%$ \\
        \bottomrule
    \end{tabular}
\end{table}

\paragraph{Both axes are causally necessary.}
Masking the action-to-retrieval cross-attention at a single decoder layer confirms that both axes carry the behavior causally rather than merely correlating with it (\cref{tab:pusht-mech-ablation}). We set the action$\to$\{ret-1, ret-end, retrieved proprio, retrieved action\} attention to $-\infty$ before softmax, independently at L10 and at L15, across all seven angles. Blocking L10 intake on the five-success rollouts drops success by $20$--$80$ percentage points, so reading the retrieved plan is required to succeed; blocking L15 commit drops it by $20$--$60$ pp, so retrieved evidence alone is not enough without the commitment step. The same L10 mask is equally telling on failures: applied to the five-failure rollouts it \emph{recovers} nontrivial success on 6 of 7 angles, from $0\%$ to $20$--$40\%$, with the largest rescue at $0^\circ$ and $\pm 60^\circ$, the very angles that over-anchor on retrieval (\cref{tab:pusht-mech-l10}). Excessive L10 intake is therefore not a benign side effect but a direct cause of failure, which sets up the failure analysis in \cref{app:pusht-failure}.

\section{PushT Failure Case Analysis}
\label{app:pusht-failure}

\begin{figure}[t]
    \centering
    \includegraphics[width=0.95\linewidth]{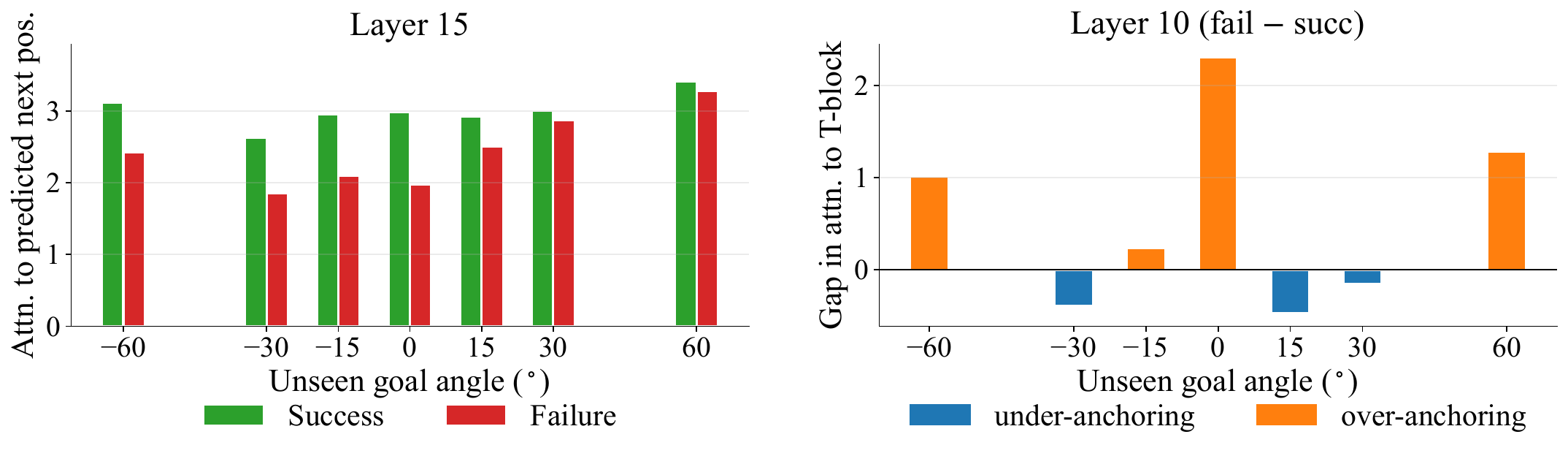}
    \caption{Two failure modes across all seven unseen angles. (a) L15 own-prediction commit, success vs failure: success bars exceed failure bars at every angle, so weakened L15 commit is a universal failure signature, with the strongest weakening at $0^\circ$. (b) L10 retrieval-intake gap, failure $-$ success: angles close to the seen training band ($\pm 30^\circ$, $+15^\circ$) show negative gaps (\emph{under-anchoring}: failures under-use retrieval), while angles far from it ($0^\circ$, $\pm 60^\circ$) show positive gaps (\emph{over-anchoring}: failures over-rely on retrieval); $0^\circ$ is the heaviest over-anchoring.}
    \label{fig:pusht-failure-modes}
\end{figure}

This section follows the two axes from \cref{app:pusht-mechanism} into the failure regime and finds that each one breaks in its own way: L15 commit weakens on essentially every failure, whereas L10 intake is distorted in a direction that depends on the goal angle. For each unseen angle we split the ten probe rollouts into a five-success and a five-failure subset and compare the axis-specific attention measurement between them (\cref{fig:pusht-failure-modes}), which isolates one failure signature per axis and keeps the diagnosis aligned with the mechanism established above.

\begin{table}[t]
    \centering
    \caption{\textbf{L15 commit weakening on failures.} Action-slot attention from the last retrieved frame to the policy's predicted end-of-chunk pose at L15, mean over $5$ episodes per cell.}
    \label{tab:pusht-mech-l15}
    \small
    \begin{tabular}{lccc}
        \toprule
        Angle ($^\circ$) & Success & Failure & $\Delta$ (succ $-$ fail) \\
        \midrule
        $-60$ & $3.12$ & $2.43$ & $+0.69$ \\
        $-30$ & $2.63$ & $1.85$ & $+0.78$ \\
        $-15$ & $2.95$ & $2.09$ & $+0.86$ \\
        $0$   & $2.99$ & $1.97$ & $\mathbf{+1.01}$ \\
        $+15$ & $2.92$ & $2.50$ & $+0.42$ \\
        $+30$ & $3.01$ & $2.87$ & $+0.14$ \\
        $+60$ & $3.41$ & $3.28$ & $+0.14$ \\
        \bottomrule
    \end{tabular}
\end{table}

\paragraph{Axis 1 (L15 commit): weakened on every failure.}
The L15 commit axis weakens on failure at all seven unseen angles, which makes it the universal, condition-independent failure signature. At every angle the action slot attends less to its predicted end-of-chunk pose on failure than on matched success rollouts (\cref{fig:pusht-failure-modes}a, \cref{tab:pusht-mech-l15}): the success-minus-failure gap is positive everywhere, growing from $+0.14$ at the easy $+30^\circ$ and $+60^\circ$ angles to $+1.01$ at the hardest angle, $0^\circ$. Because L15 commit is the step at which the policy converts retrieved evidence and the current observation into the action it will execute, a weaker L15 signal means the policy has failed to commit to a correct action under that unseen condition, independently of how well it read the retrieval at L10. Crucially, the sign of the effect never flips across angles, so L15 commit behaves as a single axis that can be addressed condition-independently.

\begin{table}[t]
    \centering
    \caption{\textbf{L10 anchoring spectrum on failures.} Action-slot attention from the first retrieved frame to the current T-block region at L10, mean over $5$ episodes per cell. Negative $\Delta$ (fail $-$ succ) is under-anchoring; positive is over-anchoring.}
    \label{tab:pusht-mech-l10}
    \small
    \begin{tabular}{lcccl}
        \toprule
        Angle ($^\circ$) & Success & Failure & $\Delta$ (fail $-$ succ) & Zone \\
        \midrule
        $+15$ & $5.33$ & $4.85$ & $-0.47$ & Under \\
        $-30$ & $5.58$ & $5.19$ & $-0.39$ & Under \\
        $+30$ & $5.73$ & $5.58$ & $-0.15$ & Under \\
        $-15$ & $4.89$ & $5.13$ & $+0.24$ & Mild over \\
        $-60$ & $5.02$ & $6.03$ & $+1.02$ & Moderate over \\
        $+60$ & $4.32$ & $5.61$ & $+1.28$ & Moderate over \\
        $0$   & $4.32$ & $6.62$ & $\mathbf{+2.31}$ & Heavy over \\
        \bottomrule
    \end{tabular}
\end{table}

\paragraph{Axis 2 (L10 intake): an under-/over-anchoring spectrum.}
The L10 intake axis fails in opposite directions depending on how far the goal angle sits from the seen training band, so unlike L15 it is condition-specific in both magnitude and sign. Near a seen angle ($\pm 30^\circ$, $+15^\circ$), failures \emph{under-anchor}: they take in less retrieval than matched successes (gaps of $-0.15$ to $-0.47$ in \cref{tab:pusht-mech-l10}, \cref{fig:pusht-failure-modes}b), because the rollout resembles a condition the policy already handles and the failure mode is to neglect the retrieved evidence. Far from the seen band ($0^\circ$, $\pm 60^\circ$), failures \emph{over-anchor}: they take in more retrieval (gaps up to $+2.31$ at $0^\circ$, over twice the next largest), because the unfamiliar condition needs the retrieved plan more and the failure mode is to over-rely on it, staying at intake and never committing at L15. The crossover from under- to over-anchoring tracks distance from $\pm 45^\circ$ monotonically, so a single axis explains both ends of the spectrum, with $0^\circ$ as the extreme over-anchoring case and the same angle at which L15 commit is weakest.

\paragraph{A behavior-level split mirrors the attention axes.}
Independently of attention, every failure rollout falls into one of two behavioral clusters, and the dominant one is the behavioral counterpart of the attention failures above. We label rollouts by coverage, the fraction of T-block area moved toward the goal: \textbf{over-anchored} failures (coverage $\ge 0.10$) move the block but never complete the push, while \textbf{non-engaging} failures (coverage $<0.10$) leave the policy largely inactive. Across the seven unseen angles the split is $25$ over-anchored to $10$ non-engaging ($71\%$ vs.\ $29\%$); both clusters occur at every angle, and over-anchored dominates except at $-30^\circ$ ($2$ vs.\ $3$). The over-anchored cluster is what L10 over-anchoring and weakened L15 commit look like in behavior, namely a policy that keeps adjusting but never decisively executes, whereas the non-engaging cluster reflects a degenerate collapse of the policy's own prediction that reshaping retrieval attention cannot fix.

\paragraph{Implications for future work.}
The two axes call for different fixes because one fails universally and the other fails condition-specifically. L15 commit weakens in the same direction at every angle, so it can be targeted directly, for example by supervising the action-to-predicted-end attention at training time to strengthen commitment under unseen conditions. L10 intake fails in opposite directions across angles, so it instead needs adaptive control, for example an attention temperature or gate that curbs over-anchoring at hard angles while preserving engagement at easy ones. Neither remedy touches the non-engaging cluster, which lies outside both axes and is better treated as a separate robustness problem for the policy's own-prediction pathway.

\section{RoboTwin Setup Details}
\label{app:robotwin-setup}

\paragraph{Train and test tasks.}
The policy is fine-tuned on five Aloha-Agilex (target) tasks paired with UR5 (pool) retrievals: \textit{Place Cans Plasticbox} (50 paired episodes), \textit{Move Can Pot} (48), \textit{Open Microwave} (49), \textit{Grab Roller} (49), and \textit{Pick Dual Bottles} (50), for 246 paired training episodes in total. Cross-task evaluation uses five held-out tasks unseen during fine-tuning: \textit{Move Pillbottle Pad}, \textit{Lift Pot}, \textit{Click Bell}, \textit{Hand-over Mic}, and \textit{Place Bread Skillet}. A representative paired observation is shown in \cref{fig:robotwin-example} of the main body.

\paragraph{Excluded query episodes.}
A small number of query episodes are removed from the training set because their cross-embodiment retrieval is catastrophically mis-aligned and would inject systematically wrong supervision: \textit{Move Can Pot} episodes 32 and 45 (mid-task rotation mismatch with visual and quaternion-cosine outliers), \textit{Open Microwave} episode 34 (75-frame plateau with the retrieval stuck), and \textit{Grab Roller} episode 38 (quaternion hemisphere wrap-around).

\paragraph{Progressive retrieval pool.}
The test-time pool-progression eval grows the retrievable pool through five strictly nested levels, from $11$ to $35$ UR5 tasks, with the policy frozen throughout (\cref{tab:robotwin-pool}). The base of every level is the five training tasks (under the \texttt{ur5\_clean\_50} and \texttt{ur5\_randomized\_500} splits); each level then appends six further UR5 tasks in a fixed order, and exactly one of the five held-out evaluation tasks (in bold) becomes retrievable at each level, which is what drives the per-level success increase in \cref{fig:robotwin-progression}.

\begin{table}[t]
    \centering
    \caption{\textbf{Progressive retrieval pool composition (RoboTwin).} Each level is a strict superset of the previous one and appends six UR5 tasks; the held-out evaluation task that becomes retrievable at that level is in \textbf{bold}. The base is the five training tasks under the \texttt{ur5\_clean\_50} and \texttt{ur5\_randomized\_500} splits.}
    \label{tab:robotwin-pool}
    \small
    \begin{tabular}{@{}l p{0.74\linewidth}@{}}
        \toprule
        Level (size) & UR5 tasks added (held-out eval task in \textbf{bold}) \\
        \midrule
        Base (5)    & Place Cans Plasticbox, Move Can Pot, Open Microwave, Grab Roller, Pick Dual Bottles \\
        \addlinespace
        Pool 2 (11) & \textbf{Lift Pot}, Place Bread Basket, Beat Block Hammer, Move Stapler Pad, Press Stapler, Click Alarmclock \\
        \addlinespace
        Pool 3 (17) & \textbf{Hand-over Mic}, Dump Bin Bigbin, Adjust Bottle, Open Laptop, Place Mouse Pad, Shake Bottle \\
        \addlinespace
        Pool 4 (23) & \textbf{Move Pillbottle Pad}, Place Burger Fries, Place Phone Stand, Place Dual Shoes, Place Object Basket, Move Playingcard Away \\
        \addlinespace
        Pool 5 (29) & \textbf{Place Bread Skillet}, Place Container Plate, Place Fan, Place Object Scale, Hanging Mug, Place Can Basket \\
        \addlinespace
        Pool 6 (35) & \textbf{Click Bell}, Stack Bowls, Hand-over Block, Turn Switch, Stamp Seal, Place Empty Cup, Rotate QR Code \\
        \bottomrule
    \end{tabular}
\end{table}

\paragraph{Retrieval scoring weights.}
Proprioception is encoded as a sign-flip-invariant 20-D representation (xyz position $+$ 6-D rotation $+$ gripper, with sign-flips on the rotation hemisphere collapsed). Position is up-weighted ($w_{\text{pos}}=4.0$) against the 6-D rotation component. The subframe-matching cost weights are: proprio $1.0$, proprio-history (8-step window) $0.05$, action chunk $0.1$, delta-action $0.1$, image $0.1$, time $0.0$. The trajectory-level prefilter keeps the top-$5$ candidate trajectories by initial-scene similarity with a SAM3 width-height term of weight $1.0$.

\section{Retrieval Rule Details}
\label{app:retrieval-detail}

The main text (\cref{sec:method-retrieval}) states the retrieval rule in abbreviated form. This section gives the full procedure. Recall from \cref{sec:problem} that at each control step $t$ the policy selects a pool subframe $t'$ whose surrounding chunk conditions the prediction. Searching every subframe of every pool trajectory under the full feature distance at every step does not scale, so we factor the rule into two stages: a trajectory-level prefilter (Stage 1) that narrows $\mathcal{D}^{\text{pool}}$ to a small candidate set, and a subframe-level match (Stage 2) that picks the moment within those candidates whose local state best matches the query. The same two stages run at training and at deployment; only the active pool $\mathcal{D}^{\text{pool}}$ and the subframe cost differ between the two, as detailed below.

\paragraph{Stage 1: initial-scene trajectory retrieval.}
The first stage narrows the pool to trajectories whose \emph{initial scene} matches the query's. We compare only the first frame of each pool trajectory $\tau \in \mathcal{D}^{\text{pool}}$ against the query's first frame on three signals: a language embedding of the goal instruction, the initial positions of task-relevant objects extracted via SAM3~\citep{carion2025sam3segmentconcepts}, and the initial proprioception. Writing $\psi_0(\cdot)$ for this composite initial-scene descriptor, we keep the top-$K$ closest trajectories,
\begin{equation}
    \mathcal{C}^{\text{traj}}_t \;=\; \operatorname*{Top\text{-}K}_{\tau \in \mathcal{D}^{\text{pool}}} \; \Bigl(\,-\bigl\lVert \psi_0(\text{query}) - \psi_0(\tau) \bigr\rVert^2\,\Bigr),
    \label{eq:traj-retrieval}
\end{equation}
where $\operatorname{Top\text{-}K}$ returns the $K$ trajectories with the smallest distance. This stage plays two roles. It speeds up inference, since Stage 2 runs at every control step and shrinking its search set from the entire pool to $K$ trajectories is what makes step-wise retrieval feasible on large pools. It also improves precision, since trajectories with the wrong instruction, object layout, or initial arm pose are filtered out before any subframe is scored, so Stage 2 does not spend its budget comparing the query against semantically irrelevant candidates whose local frames may happen to look similar by accident.

\paragraph{Stage 2: subframe retrieval (training).}
Within $\mathcal{C}^{\text{traj}}_t$, we locate the subframe whose local state best matches the query's. A subframe is described by four heterogeneous features: the task-relevant object pose $\phi_{\text{obj}}$ (T-block position in PushT, SAM3-anchored object poses in RoboTwin), a DINO image feature $\phi_{\text{vis}}$~\citep{simeoni2025dinov3} of the current scene, the proprioception $\phi_{\text{prop}}$, and (at training only) the upcoming action chunk $\phi_{\text{act}}$. These features live on different scales and admit different natural distances, so we score them with a weighted combination rather than a single Euclidean norm,
\begin{equation}
\begin{aligned}
    d_{\text{tr}}\bigl(t,\, t'\bigr) \;=\;
    &w_{\text{obj}}\, d_{L_2}\!\bigl(\phi_{\text{obj}}^{(t)},\, \phi_{\text{obj}}^{(t')}\bigr)
    + w_{\text{prop}}\, d_{L_2}\!\bigl(\phi_{\text{prop}}^{(t)},\, \phi_{\text{prop}}^{(t')}\bigr) \\
    + &w_{\text{vis}}\, d_{\cos}\!\bigl(\phi_{\text{vis}}^{(t)},\, \phi_{\text{vis}}^{(t')}\bigr)
    + w_{\text{act}}\, d_{L_2}\!\bigl(\phi_{\text{act}}^{(t)},\, \phi_{\text{act}}^{(t')}\bigr),
\end{aligned}
\label{eq:cost-train}
\end{equation}
where $\phi_{\bullet}^{(t)}$ abbreviates the corresponding feature of the query subframe at time $t$ (analogously for the pool side at $t'$). The retrieved index is
\begin{equation}
    t' \;=\; \arg\min_{t'}\; d_{\text{tr}}(t,\, t'),
    \qquad (s_{t'}^{\text{pool}},\, a_{t'}^{\text{pool}}) \in \mathcal{C}^{\text{traj}}_t,
    \label{eq:argmin-train}
\end{equation}
and the chunk $(s_{t':t'+H}^{\text{pool}}, a_{t':t'+H-1}^{\text{pool}})$ starting at $t'$ becomes the retrieval conditioning of \cref{eq:policy}.

\paragraph{Distance choices.}
We use squared $L_2$ distance $d_{L_2}$ for the low-dimensional geometric features (object pose $\phi_{\text{obj}}$, proprioception $\phi_{\text{prop}}$, action chunk $\phi_{\text{act}}$) and cosine distance $d_{\cos}$ for the high-dimensional DINO embedding $\phi_{\text{vis}}$. The visual weight $w_{\text{vis}}$ is kept small in practice, so DINO acts as a soft visual sanity check rather than dominating the geometric components, which carry the metric signal the retrieved chunk is meant to align with.

\paragraph{Role of the action term at training.}
The action term $\phi_{\text{act}}$ is the one feature available at training but not at deployment, so its role is worth stating explicitly. Including it in \cref{eq:cost-train} means the retrieved chunk matches not only the local state but also what the demonstration was about to do, which gives the WAM a tightly aligned training signal: the retrieved chunk is what a trajectory close to the query's own actually executed next. The residual in \cref{eq:residual} therefore has to encode only a small embodiment-specific correction, and the next-state objective supplies the dense signal to learn it.

\paragraph{Stage 2: subframe retrieval (inference).}
At deployment the future query action $a_{t:t+H-1}^{\text{query}}$ is unknown, since it is exactly what the WAM is predicting, so the subframe cost drops the action term while keeping the Stage 1 trajectory filter \cref{eq:traj-retrieval} unchanged,
\begin{equation}
    d_{\text{inf}}(t,\, t') \;=\;
    w_{\text{obj}}\, d_{L_2}\!\bigl(\phi_{\text{obj}}^{(t)},\, \phi_{\text{obj}}^{(t')}\bigr)
    + w_{\text{prop}}\, d_{L_2}\!\bigl(\phi_{\text{prop}}^{(t)},\, \phi_{\text{prop}}^{(t')}\bigr)
    + w_{\text{vis}}\, d_{\cos}\!\bigl(\phi_{\text{vis}}^{(t)},\, \phi_{\text{vis}}^{(t')}\bigr),
    \label{eq:cost-infer}
\end{equation}
with the retrieved index $t' = \arg\min_{t'} d_{\text{inf}}(t,\, t')$ over $\mathcal{C}^{\text{traj}}_t$. The model is trained against the richer state-and-action match of \cref{eq:cost-train}, so it absorbs the train-to-inference mismatch into the action latents rather than requiring the inference-time match to reproduce the action term.

\paragraph{Pool extension at deployment.}
The operating mode this rule supports is that, at deployment, the practitioner may bring \emph{new} pool-embodiment demonstrations for tasks the model has never seen during training. These demonstrations replace the active pool, $\mathcal{D}^{\text{pool}} \leftarrow \mathcal{D}_{\text{test}}^{\text{pool}}$, and are reindexed under $\psi_0$ for Stage 1 \cref{eq:traj-retrieval} and under $(\phi_{\text{obj}}, \phi_{\text{prop}}, \phi_{\text{vis}})$ for Stage 2 \cref{eq:cost-infer}. This pool extension touches no model parameters, so new tasks are added by indexing new data, not by retraining (the property tested in \cref{fig:pusht-progression,fig:robotwin-progression}). Because both stages re-run against the current index, the practitioner can also grow $\mathcal{D}_{\text{test}}^{\text{pool}}$ within a session and see coverage expand without restart.

\paragraph{Per-feature weights.}
The feature weights $(w_{\text{obj}}, w_{\text{prop}}, w_{\text{vis}}, w_{\text{act}})$ and the proprioception encoding are benchmark-specific. The RoboTwin settings, including the sign-flip-invariant 20-D proprioception representation, the per-component subframe weights, and the top-$5$ trajectory prefilter, are listed in \cref{app:robotwin-setup}. On PushT, $\phi_{\text{obj}}$ is the T-block pose, the geometric terms dominate, and $w_{\text{vis}}$ is set low so that DINO only breaks ties between geometrically comparable subframes.

\paragraph{Inference loop.}
\begin{algorithm}[t]
\caption{Test-time inference with retrieval (one task episode).}
\label{alg:inference}
\begin{algorithmic}[1]
\Require trained WAM $\pi_\theta$; pool $\mathcal{D}_{\text{test}}^{\text{pool}}$ pre-indexed for trajectory retrieval (\cref{eq:traj-retrieval}); per-component features $(\phi_{\text{obj}}, \phi_{\text{prop}}, \phi_{\text{vis}})$; horizon $H$; action stride $K \le H$.
\State $t \gets 0$
\State $\mathcal{C}^{\text{traj}} \gets$ Top-$K$ trajectories in $\mathcal{D}_{\text{test}}^{\text{pool}}$ under $\psi_0$ distance to the query (\cref{eq:traj-retrieval})
\While{episode not terminated}
    \State observe query state $s_t^{\text{query}}$
    \State \textbf{subframe retrieval:}\;
        $t' \gets \arg\min_{t'}\, d_{\text{inf}}(t,\, t')$ over $\mathcal{C}^{\text{traj}}$
        \Comment{inference cost \cref{eq:cost-infer}}
    \State fetch retrieved chunk $\mathbf{r}_t \gets (s_{t':t'+H}^{\text{pool}},\, a_{t':t'+H-1}^{\text{pool}})$
    \State \textbf{predict:}\;
        $\hat a_{t:t+H-1}^{\text{query}} \gets \pi_\theta\!\bigl(s_t^{\text{query}},\, \mathbf{r}_t\bigr)$
        \Comment{residual: $\hat a = a^{\text{pool}} + \Delta a$ (\cref{eq:residual})}
    \State execute first $K$ actions $\hat a_{t:t+K-1}^{\text{query}}$
    \State $t \gets t + K$
\EndWhile
\end{algorithmic}
\end{algorithm}
\cref{alg:inference} assembles the two stages into the full test-time loop for one episode. Stage 1 runs once at the start, fixing the candidate set $\mathcal{C}^{\text{traj}}$ from the initial scene, while Stage 2 re-runs every control step: the policy observes the current query state, selects the matching pool subframe $t'$ under the inference cost \cref{eq:cost-infer}, fetches the chunk starting at $t'$, and predicts the residual action chunk \cref{eq:residual}. It then executes the first $K$ of the $H$ predicted actions and advances by $K$, so the model re-retrieves and re-plans as the scene evolves rather than committing to a single retrieved trajectory for the whole episode.

\section{Hyperparameters}
\label{app:hyperparams}

This section collects the backbone, retrieval, and training hyperparameters in one place (\cref{tab:hyperparams}). The retrieval values are the RoboTwin configuration; on PushT the object feature $\phi_{\text{obj}}$ is the T-block pose, the geometric terms dominate, and the visual weight is kept low, as described qualitatively in \cref{app:retrieval-detail}.

\begin{table}[t]
    \centering
    \caption{\textbf{Hyperparameters.} Backbone, and the RoboTwin 2.0 retrieval and training settings. The Stage 2 weights are the per-component subframe-matching weights of \cref{eq:cost-train}, and the proprioception encoding feeds $\phi_{\text{prop}}$.}
    \label{tab:hyperparams}
    \small
    \begin{tabular}{@{}ll@{}}
        \toprule
        Hyperparameter & Value \\
        \midrule
        \multicolumn{2}{@{}l}{\textit{Backbone}} \\
        Base video model            & Cosmos Predict2.5 (2B) \\
        Output                      & joint action + future image (WAM), flow matching \\
        Action parameterization     & residual over the retrieved chunk \\
        \midrule
        \multicolumn{2}{@{}l}{\textit{Retrieval, Stage 1 (trajectory prefilter)}} \\
        Top-$K$ trajectories        & $5$ \\
        SAM3 width-height term weight & $1.0$ \\
        \midrule
        \multicolumn{2}{@{}l}{\textit{Retrieval, Stage 2 (subframe scoring weights)}} \\
        Proprioception $w_{\text{prop}}$         & $1.0$ \\
        Proprioception history (8-step window)   & $0.05$ \\
        Image (DINO) $w_{\text{vis}}$            & $0.1$ \\
        Action chunk $w_{\text{act}}$            & $0.1$ \\
        Delta action                             & $0.1$ \\
        Time                                     & $0.0$ \\
        \midrule
        \multicolumn{2}{@{}l}{\textit{Retrieval, proprioception encoding}} \\
        Representation              & 20-D, 6-D rotation (sign-flip invariant) \\
        Position weight $w_{\text{pos}}$         & $4.0$ \\
        \midrule
        \multicolumn{2}{@{}l}{\textit{Training}} \\
        Retrieval pairs per sample (top-$k$)     & $3$ (weighted, top-$N=5$ prefilter) \\
        Horizon $H$ (chunk size)                 & $16$ frames \\
        Action stride $K$                        & $1$ \\
        Action-loss weight $\lambda$             & $16$ \\
        Optimizer                                & AdamW \\
        Learning rate                            & $1\times10^{-4}$ (cosine, $2000$-step warm-up) \\
        Batch size                               & $16$/GPU $\times\,4$ grad.\ accum.\ $\times\,8$ GPUs $=512$ \\
        Training iterations                      & $20{,}000$ \\
        \bottomrule
    \end{tabular}
\end{table}

\end{document}